\documentclass[runningheads]{llncs}
\usepackage{etex}
\usepackage{hyperref}
\usepackage[utf8]{inputenc}
\usepackage[T1]{fontenc}
\usepackage[ngerman,english]{babel}
\usepackage{wrapfig}
\usepackage{times}
\usepackage{paralist}
\usepackage{graphicx}
\usepackage[dvipsnames]{xcolor}
\usepackage{subfig}
\usepackage{booktabs}
\usepackage{caption}
\usepackage{paralist}
\usepackage{amssymb, amsfonts}
\usepackage{amsmath}
\usepackage{amsopn}
\usepackage{url}
\usepackage{multirow}
\usepackage{relsize}
\usepackage{caption}
\usepackage{xfrac}
\usepackage{pdflscape}
\usepackage{cite}
\usepackage{tabularx}
\newcommand{\mypar}[1]{\smallskip\noindent\textbf{#1.}}
\newcommand{\mypartwo}[1]{\smallskip\noindent\textit{#1.}}

\newcommand{\mytitle}{
		Extracting Semantic Process Information from the Natural Language in Event Logs
}

\hypersetup{
	pdftoolbar=true,        
	pdfmenubar=true,        
	pdffitwindow=false,     
	pdfstartview={FitH},    
	pdftitle={\mytitle},    
	pdfauthor={},     
	pdfsubject={},   
	pdfcreator={},   
	pdfproducer={}, 
	pdfkeywords={}, 
	pdfnewwindow=true,      
	colorlinks=true,       
	linkcolor=Brown,          
	citecolor=OliveGreen,        
	filecolor=magenta,      
	urlcolor=NavyBlue           
}

\DeclareMathAlphabet\mathbfcal{OMS}{cmsy}{b}{n}

\newcommand{\objectname}{\texttt{obj}}

\newcommand{\objectstatus}{\texttt{obj$_{status}$}}
\newcommand{\objectstatusshort}{\texttt{obj$_{s}$}}
\newcommand{\actionname}{\texttt{action}}

\newcommand{\actionstatus}{\texttt{action$_{status}$}}
\newcommand{\actionstatusshort}{\texttt{action$_{s}$}}
\newcommand{\actorname}{\texttt{actor}}
\newcommand{\actorinstance}{\texttt{actor$_{instance}$}}

\newcommand{\passivename}{\texttt{passive}}
\newcommand{\passiveinstance}{\texttt{passive$_{instance}$}}

\newcommand{\attributes}{\mathcal{D}}
\newcommand{\textattributes}{\mathcal{D}^T}
\newcommand{\miscattributes}{\mathcal{D}^M}
\newcommand{\classifierattributes}{\mathcal{D}^\mathcal{L}}

\newcommand{\labelroles}{\mathcal{R}'}

\begin{document}
\title{\mytitle}
\titlerunning{Extracting Semantic Process Information from Event Logs}
%
\author{Adrian Rebmann \and Han van der Aa}
\authorrunning{A. Rebmann \and H. van der Aa}
%
\institute{Data and Web Science Group, University of Mannheim, Mannheim, Germany
\email{\{rebmann|han\}@informatik.uni-mannheim.de}\\
}
\maketitle              
\begin{abstract}
Process mining focuses on the analysis of recorded event data in order to gain insights about the true execution of business processes. 
While foundational process mining techniques treat such data as sequences of abstract events, more advanced techniques depend on the availability of specific kinds of information, such as resources in organizational mining and business objects in artifact-centric analysis. 
However, this information is generally not readily available, but rather associated with events in an ad hoc manner, often even as part of unstructured textual attributes. 
Given the size and complexity of event logs, this calls for automated support to extract such  process information and, thereby, enable advanced process mining techniques. 
In this paper, we present an approach that achieves this through so-called semantic role labeling of event data. We combine the analysis of textual attribute values, based on a state-of-the-art language model, with a novel attribute classification technique. 
In this manner, our approach extracts information about up to eight semantic roles per event.  We demonstrate the approach's efficacy through a quantitative evaluation using a broad range of event logs and demonstrate the usefulness of the extracted information in a case study.

\end{abstract}
\section{Introduction}
\label{sec:introduction}

Process mining~\cite{vanderaalst2016data} enables the analysis of business processes based on event logs that are recorded by information systems in order to gain insights into how processes are truly executed.
Process mining techniques obtain these insights by analyzing sequences of recorded events, also referred to as \emph{traces}, that jointly comprise an \emph{event log}. 
Most foundational process mining techniques treat traces as sequences of abstract symbols, e.g., $\langle a, b, c, d \rangle$. However, more advanced techniques, such as social network analysis~\cite{vanderaalst2005discovering} and object-centric process discovery~\cite{VanderAalst2019} 
go beyond this abstract view and consider specific kinds of information contained in the events' labels or attributes, such as \emph{actors}, \emph{business objects}, and  \emph{actions}.

A key inhibitor of such advanced process mining techniques is that the required pieces of information, which we shall refer to as \emph{semantic components}, are not readily available in most event logs. 
A prime cause for this is the lack of standardization of attributes in event logs. 
While the XES standard~\cite{xes} defines certain standard extensions for attributes (e.g.,  \texttt{org:resource}), the use of these conventions is not enforced and, thus, not necessarily followed by real-life logs (cf.,~\cite{bpi14}). Furthermore, the standard only covers a limited set of attributes, which means that  information on components such as \emph{actions} and \emph{business objects}, are not covered by the standard at all and, therefore, often not explicitly represented in event logs. 

Rather, relevant information is often captured as part of unstructured, textual data attributes associated with events, most commonly in the form of an event's label. For example, the ``\emph{Declaration submitted by supervisor}'' label from the most recent BPI Challenge~\cite{bpi20} captures information on the business object (\emph{declaration}), the action (\emph{submitted}), and the actor (\emph{supervisor}). 
Since these components are all encompassed within a single, unstructured text, the information from the label cannot be exploited by process mining techniques.
Enabling this use, thus, requires the processing of each individual attribute value in order to extract the included semantic information. 
Clearly, this is an extremely tedious and time-consuming task when considered in light of 
the complexity of real-life logs, with hundreds of event classes, dozens of attributes, and thousands of instances. Therefore, this calls for automated support to extract semantic components from event data and make them available to process mining techniques.

To achieve this, we propose an approach that automatically extracts semantic information from events while imposing no assumptions on a log's attributes.
In particular, it aims to extract information on eight \emph{semantic roles}, covering various kinds of information related to business objects, actions, actors, and other resources. 
The choice for these specific roles is based on their relevance to existing process mining techniques and presence in available real-life event logs. To achieve its goal, our approach combines state-of-the-art natural language processing (NLP) techniques, tailored to the task of semantic role labeling, with a novel technique for semantic attribute classification.

Following an illustration of the addressed problem (\autoref{sec:motivation}) and presentation of our approach itself (\autoref{sec:approach}), the quantitative evaluation presented in \autoref{sec:evaluation} demonstrates that our approach achieves accurate results on real-life event logs, spanning various domains and varying considerably in terms of their informational structure.  
Afterwards, \autoref{sec:application} highlights the usefulness of our approach by using it to analyze an event log from the 2020 BPI Challenge (BPI20).
Finally, \autoref{sec:relatedwork} discusses streams of related work, before concluding in \autoref{sec:conclusion}.

\section{Motivation}
\label{sec:motivation}

This section motivates the goal of semantic role labeling of event data (\autoref{sec:annotationgoal}) and discusses the primary challenges associated with this task (\autoref{sec:annotationchallenges}).

\subsection{Semantic Roles in  Event Data}
	\label{sec:annotationgoal}
	
	Given an event log, our work sets out to label pieces of information associated with events that correspond to particular \emph{semantic roles}. In this work, we focus on various roles that support a detailed analysis of business process execution from a behavioral perspective, i.e., we target semantic roles that are commonly observed in event logs and that are relevant for an order-based analysis of event data.	
	Therefore, we consider information related to four main categories: \emph{business objects}, \emph{actions}, as well as \emph{active} and \emph{passive} resources involved in a process' execution.
	For each category, we define multiple semantic roles, which we jointly capture in a set $\mathcal{R}$:
		
	\mypar{Business objects}  In line with convention~\cite{mendling2010activity}, we use the term \emph{business object} to broadly refer to the main object(s) relevant to an event. 
	Particularly, we define (1) \objectname\ as 
	 the type of business object to which an event relates, e.g.,  a \emph{purchase order}, an \emph{applicant}, or a \emph{request} and (2) \objectstatus\ as an object's status, e.g., \emph{open}  or \emph{completed}.

	\mypar{Actions}
	We define two roles to capture information on the actions that are applied to business objects : (1) \actionname, as the kind of action, e.g., \emph{create}, \emph{analyze}, or \emph{send}, and (2) \actionstatus, as further information on its status, e.g., \emph{started}  or \emph{paused}.

	\mypar{Actors}
		Information regarding the active resource in the event is captured in the following two roles: (1) \actorname\ as the type of active resource in the event, e.g., a ``\emph{supervisor}'' or a ``\emph{system}'', and (2) \actorinstance\ for information indicating the specific actor instance, e.g., an employee identifier.

	\mypar{Passive resources}
	Aside from the actor, events may also store information on \emph{passive} resources involved in an event, primarily in the form of \emph{recipients}. For this, we again define two roles: (1) \passivename\ as the type of passive resource related to the event, e.g., the role of an employee receiving a document or a system on which a file is stored or transferred through, and (2) \passiveinstance\ for information indicating the specific resource, e.g., an employee or system identifier.

\smallskip
The considered semantic roles enable a broad range of fine-granular insights into the execution of a process. For example, the \emph{business object} and \emph{action} categories allow one to obtain detailed insights into the business objects moving through a process, their inter-relations, and their life-cycles. Furthermore, by also considering the resource-related roles, one can, for instance, gain detailed insights into the resource behavior associated with a particular business object, e.g., how resources jointly collaborate on the processing of a specific document.
While the covered roles, thus, support a wide range of analyses and are purposefully selected based on their relevance in real-life event logs, our approach is by no means limited to these specific roles. Given that we employ state-of-the-art NLP technology that generalizes well, the availability of appropriate event data allows our approach to be easily extended to cover additional semantic roles, both within and outside the informational categories considered here.

\subsection{The Semantic Role Labeling Task}
	\label{sec:annotationchallenges}
	
To ensure that all relevant information is extracted from an event log, our work considers two aspects of the \emph{semantic role labeling} task, concerned with two kinds of event attributes: 
\emph{attribute-level classification} for attributes dedicated to a single semantic role and \emph{instance-level labeling} for textual attributes covering various roles:

\mypar{Attribute-level classification}
Attribute-level classification sets out to determine the role of attributes that correspond to the same, dedicated semantic role throughout an event log, e.g., a \texttt{doctype} attribute indicating a business object. Although the XES standard~\cite{xes} specifies several standard event attributes, such as \texttt{org:resource} and \texttt{org:role}, these only cover a subset of the semantic roles we aim to identify. They omit roles related to business objects, actions, and passive resources. 
These other semantic roles may, thus, be captured in attributes with diverse names, e.g., the \objectstatus\ role corresponds to event attributes such as  \texttt{isClosed} or  \texttt{isCancelled} in the Hospital log\footnote{We kindly refer to \autoref{sec:evaluationdata} for further information on the event logs referenced here.}.
Furthermore, even for roles covered by standard attributes, there is no guarantee that event logs adhere to the conventions, e.g., rather than using \texttt{org:group}, the BPI14 log captures information  on actors in an \texttt{Assignment\_Group} attribute.

\mypar{Instance-level labeling}
Instance-level labeling, instead, sets out to derive semantic information from attributes with unstructured, textual values that encompass various semantic roles, differing per event instance. This task is most relevant for so-called event labels, often stored in a \texttt{concept:name} attribute. These labels contain highly valuable semantic information, yet also present considerable challenges to their proper handling, as illustrated through the real-life event labels in \autoref{tab:instancelevelexamples}.
The examples highlight the diversity of textual labels, in terms of their structure and the semantic roles that they cover.
It is worth mentioning that such differences may even exist for labels within the same event log, e.g., labels $l_5$ and $l_6$ differ considerably in their textual structure and the information they cover, yet they both stem from the BPI19 log. 
Another characteristic to point out is the possibility of recurring roles within a label, such as seen for label $l_1$, which contains two \actionname\ components: \emph{draft} and \emph{send}. 
Hence, an approach for instance-level labeling needs to be able to deal with textual attribute values that are highly variable in terms of the information they convey, as well as their structure.

\begin{table}[!htb]
	\small
	\centering
	\begin{tabularx}{\linewidth}{lllX}
		\toprule
		\textbf{Log} & \textbf{ID}&  \textbf{Event label} & \textbf{Contained semantic roles} \\
		\midrule
		
		WABO & $l_1$ &  draft and send request for advice & \actionname~($\times$\texttt{2}), \objectname \\ 
		BPI15 & $l_2$ &send design decision to stakeholders& 
		 \actionname, \objectname, \passivename \\
		BPI15 & $l_3$ &send letter in progress& \actionname, \objectname, \actionstatus\\		
		RTFM &$l_4$ & insert date appeal to  prefecture & 
		 \actionname, \objectname, \passivename \\
		 BPI19 & $l_5$ &Vendor creates invoice & \actorname, \actionname, \objectname \\	
		 BPI19 & $l_6$ &SRM: In Transfer to Execution Syst. & \actionname, \passivename \\
		 BPI20 &$l_7$ &Declaration final\_approved by supervisor &
		 \objectname, \actionstatus, \actionname, \actorname \\ 
		\bottomrule
	\end{tabularx}
	\caption{Exemplary event labels from real-life event logs.}
	\label{tab:instancelevelexamples}
	\vspace{-2em}
\end{table}

\section{Semantic Event Log Parsing}
\label{sec:approach}

This section presents our approach for the semantic labeling of event data. Its input and main steps are as follows:

\mypar{Approach input}
Our approach takes as input an event log $L$ that consists of events recorded by an information system. We denote the universe of all events as $\mathcal{E}$, where each event $e \in \mathcal{E}$ carries information in its payload. This payload is defined by a set of (data) \emph{attributes} $\mathcal{D} = \{D_1, \ldots, D_p \}$ with $\textnormal{dom}(D_i)$ as the domain of attribute $D_i$, $1 \leq i \leq p$ and $\textnormal{name}(D_i)$, its name. We write $e.D$ for the value of $D$ for an event $e$.

Note that we do not impose any assumptions on the attributes contained in an event log $L$, meaning that we do not assume that attributes such as \texttt{concept:name} and \texttt{org:role} are included in $\mathcal{D}$.

\begin{figure}[!h]
	\centering
	\includegraphics[width=\linewidth]{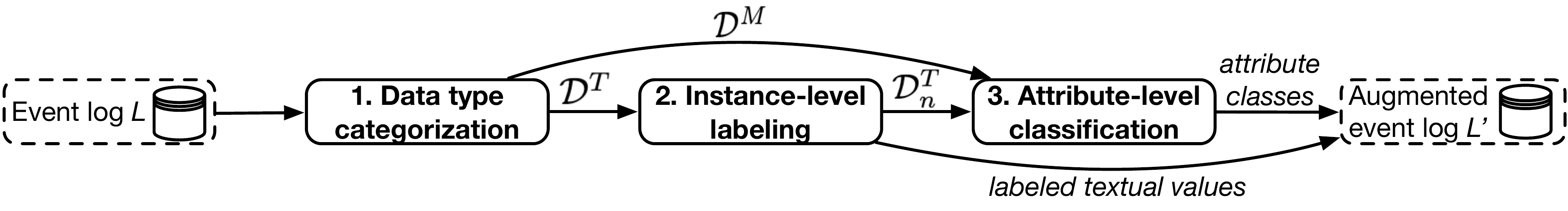}
	\caption{Overview of the approach.}
	\label{fig:approach}
\end{figure}

\mypar{Approach steps}
The goal of our approach is to label the values of event attributes with their semantic roles.
To achieve this, our approach consists of three main steps, as visualized in \autoref{fig:approach}. 
Given a log $L$ and its set of event attributes $\attributes$, Step~1 first identifies sets of \emph{textual attributes} $\textattributes \subseteq \attributes$ and of \emph{miscellaneous attributes} $\miscattributes \subseteq \attributes$.
Afterwards, Step~2 labels the values of textual attributes in $\textattributes$ to extract the parts 
that correspond to semantic roles, e.g., recognizing that a ``\emph{document received}'' event label contains the business object ``\textit{document}'' and the action ``\textit{received}''.
Step~3 focuses on the attribute-level classification of miscellaneous attributes in $\miscattributes$, as well as some textual attributes $\textattributes_n \subseteq \textattributes$ that were deemed unsuitable for instance-level labeling during the previous step. This classification step aims to determine the semantic role that corresponds to all values of a certain attribute in $\miscattributes \cup \textattributes_n$, e.g., recognizing that all values of a \texttt{doctype} attribute correspond to the \objectname\ role.

In the remainder, Sections~\ref{sec:approachstep1} through~\ref{sec:approachstep3} describe the steps of our approach in detail, whereas \autoref{sec:approachoutput} discusses how their outcomes are combined in order to obtain an event log $L'$ augmented with the extracted semantic information.

\subsection{Step 1: Data Type Categorization}
	\label{sec:approachstep1}

	In this step, our approach sets out to identify the sets of textual attributes $\textattributes$ and miscellaneous attributes $\miscattributes$. 
	As a preprocessing step, we first identify \texttt{string}, \texttt{timestamp}, and \texttt{numeric} attributes using standard libraries, e.g., 
	\emph{Pandas} in Python\footnote{https://pandas.pydata.org}. 
	
	\mypar{Identifying textual attributes}
	To identify the set of textual attributes $\textattributes$, we need to differentiate between \texttt{string} attributes with true natural language values, e.g., 
	``\emph{document received}'' or ``\emph{Create\_PurchaseOrder}'', and other kinds of alphanumeric attributes, with values such as ``\textit{A}'', ``\textit{USER\_123}'', and ``\textit{R\_45\_2A''}. 
	Only the former kind of attributes will be assigned to $\textattributes$ and, thus, analyzed on an instance-level in the remainder of the approach. We identify such  true textual attributes as follows:
	\begin{compactenum}
		\item 
		Given a \texttt{string} attribute, we first apply a tokenization function $tok$, which splits an attribute value into lowercase tokens (based on whitespace, camel-case, underscores, etc.) and omits any numeric ones. E.g., given  $s_1 =$ ``\textit{Create\_PurchaseOrder}'', $s_2=$``\textit{USER\_123}'', and $s_3=$  ``\textit{08\_AWB45\_005}'', we obtain: 
		\textit{tok}($s_1$) = [\textit{create}, \textit{purchase}, \textit{order}], \textit{tok}($s_2$) = [\textit{user}] and \textit{tok}($s_3$) = [\textit{awb}].
		
		\item We apply a \textit{part-of-speech tagger}, provided by standard NLP tools (e.g., Spacy~\cite{honnibal2017spacy}), to assign a token from the  Universal Part of Speech tag set\footnote{https://universaldependencies.org/docs/u/pos/} to each token. 
		In this manner, we obtain [(create,\emph{VERB} ) (purchase, \emph{NOUN}), (order, \emph{NOUN})] for $s_1$, [(user, \emph{NOUN})] for $s_2$,  and  [(awb, \emph{PROPN})] for $s_3$. 

		\item Finally, we exclude any attribute from $\textattributes$ that only has values with the same token in $tok(s)$ or do not contain any  \texttt{NOUN}, \texttt{VERB}, \texttt{ADV}, or \texttt{ADJ} tokens. In this way, we omit attributes with values such as $s_2=$ ``\emph{USER\_123}" and $s_3=$ ``\emph{08\_AWB45\_005}'', which are identifiers, rather than textual attributes.  The other attributes, which have diverse, textual values, e.g., $s_1=$``Create\_PurchaseOrder'', are assigned to $\textattributes$.  
	\end{compactenum}
	
	\mypar{Selecting miscellaneous attributes}
		We also identify a set of non-textual attributes that are candidates for semantic labeling, referred to as the set of miscellaneous attributes, $\miscattributes \subseteq \attributes \setminus  \textattributes$.
		This set contains attributes that are not included in $\textattributes$, yet have a data type that may still correspond to a semantic role in $\mathcal{R}$. 
		
		To achieve this, we discard those attributes in $\attributes \setminus \textattributes$ 
		categorized as \texttt{timestamp} attributes, as well as \texttt{numeric} attributes that include 
		\emph{real} or \emph{negative} values.
		We exclude these because they  are not used to capture semantic information. By contrast, the remaining attributes 
		have data types that may correspond to roles in $\mathcal{R}$, such as 
		\texttt{boolean} attributes that can be used to indicate specific states, e.g., \texttt{isClosed}, whereas non-negative integers are commonly used as identifiers. Together with the \texttt{string} attributes not selected for $\textattributes$, the retained attributes are assigned to $\miscattributes$.

\subsection{Step 2: Instance-level Labeling of Textual Attributes}
\label{sec:approachstep2}

In this step, our approach sets out to label the values of textual attributes in order to extract the parts that correspond to certain semantic roles, e.g., recognizing that a ``\emph{create purchase order}'' event label contains ``\emph{purchase order}'' as the   \objectname\ and ``\emph{create}'' as the \actionname.
 As discussed in \autoref{sec:annotationchallenges}, this comes with considerable challenges, given the high diversity of textual attribute values in terms of their linguistic structure and informational content.
 To be able to deal with these challenges, we therefore build on state-of-the-art developments in the area of natural language processing.

\mypar{Tagging task}
We approach the labeling of textual attribute values with semantic roles as a text tagging task. Therefore, we instantiate a function that assigns a semantic role to chunks (i.e., groups) of consecutive tokens from a tokenized textual attribute value. Formally,
given the tokenization of an attribute value, $tok(e.D) =  \langle t_1, \ldots, t_n\rangle $, for an attribute $D \in \textattributes$, 
we define a function $tag(\langle t_1, \ldots, t_n\rangle)  \rightarrow \langle 
c_1\textbackslash r_1, \ldots, c_m\textbackslash r_m \rangle $, where $c_i$ for $1 \leq i \leq m$ is a chunk consisting of one or more consecutive tokens from $\langle t_1, \ldots, t_n\rangle$, with $r_i \in \mathcal{R} \cup \{\texttt{other}\}$ its associated semantic role. 
For instance, $tag(\langle \textit{create}, \textit{purchase}, \textit{order} \rangle)$ yields: 
$\langle \textit{create}\textbackslash \actionname,$ $\textit{purchase order}\textbackslash \objectname \rangle$. 

 \mypar{BERT}
To instantiate the $tag$ function, we employ
BERT~\cite{devlin-etal-2019-bert}, a language model that is capable of dealing with highly diverse textual input and achieves state-of-the-art results on a wide range of NLP tasks.
BERT has been pre-trained on huge text corpora in order to develop a general understanding of a language. This model can then be \emph{fine-tuned} by training it on an additional, smaller
training data collection to target a particular task. In this manner, the trained model combines its general language understanding with aspects that are specific to the task at hand. In our case, we thus fine-tune BERT in order to tag chunks of textual attribute values that correspond to semantic roles.

\mypar{Fine-tuning}
For the fine-tuning procedure, we manually labeled a collection of 13,231 unique textual values stemming from existing collections of process models~\cite{Leopold2019}, textual process descriptions~\cite{leopold2018identifying}, and event logs (see \autoref{sec:evaluationdata}).
As expected, the collected samples do not capture information on resource instances, and rather contain information on the type level (i.e., \actorname\ and \passivename). 
For those semantic roles that are included in the samples, we observe a considerable imbalance in their commonality, as depicted in \autoref{tab:training}. 
In particular, while roles such as \objectname\ (14,629 times), \actionname\ (12,573), and even \passivename\ (1,191) are relatively common, we only found few occurrences of \actorname\ (135), \objectstatus\ (92), and \actionstatus\ (30) roles. 

\begin{table}[!htb]
	\vspace{-1.5em}
	\centering
	\caption{Training data used to fine-tune the language model, with $s=status$}
	\label{tab:training}	
	\setlength{\tabcolsep}{3.5pt}
	\begin{tabular}{lrccccccc}
		\toprule
		\textbf{Source}                       & \textbf{Count} & \textbf{\objectname} & \textbf{\objectstatusshort} & \textbf{\actionname }\  & \textbf{\actionstatusshort} &\textbf{\actorname} &\textbf{\passivename} & \textbf{\texttt{other}} 
		\\		\midrule 
		Process models               & 11,658   & 13,543    & \hphantom{1}50  & 11,445               & \hphantom{15}3            & \hphantom{2}58        & 1,058       & 4,966 \\
		
		Textual desc.        & 498      & \hphantom{13,}503      & \hphantom{1}11   & \hphantom{11,}498                & \hphantom{15}0            & \hphantom{20}8        & \hphantom{1,}114         & \hphantom{4,}206   \\
		
		Event logs                       & 625      & \hphantom{13,}583     & \hphantom{1}31    & \hphantom{11,}630                & \hphantom{1}27           & \hphantom{2}69        & \hphantom{1,0}19          & \hphantom{4,}291   \\
		
		Augmentation                 & 450      & \hphantom{13,}350     & 100    & \hphantom{11,}350                & 150          & 200       & \hphantom{1,00}0           & \hphantom{4,}150   \\
		
		\midrule
		\textbf{Total                      }  & 13,231   & 14,979  & 192       & 12,923           & 180          & 335       & 1,191       & 5,613\\
		\bottomrule 
	\end{tabular}
\end{table}

To counter this imbalance, we created additional training samples with \objectstatus, \actionstatus, and \actorname\ roles through established data augmentation strategies. In particular, we created samples by complementing randomly selected textual values with (1)  known \actorname\ descriptions, e.g., ``\emph{purchase order created}" is extended to ``\emph{purchase order created by supervisor}", and (2) common life-cycle transitions from \cite[p.131]{vanderaalst2016data} to create samples containing \objectstatus\ and \actionstatus\ roles, e.g., ``\emph{check invoice}'' is extended to ``\emph{check invoice completed}''. However, as shown in \autoref{tab:training}, we limited the number of extra samples to avoid overemphasizing the importance of these roles.

Given this training data, we operationalize the $tag$ function using the \emph{BERT base uncased pre-trained language model}\footnote{\url{https://github.com/google-research/bert}} with 12 transformer layers, a hidden state size of 768 and 12 self-attention heads. As suggested by its developers~\cite{devlin-etal-2019-bert}, we trained 2 epochs using a batch size of 16 and a learning rate of 5e-5.

\mypar{Reassigning noun-only attributes}
After applying the $tag$ function to the values of an attribute $D \in \textattributes$, we check whether the tagging is likely to have been successful. In particular, we recognize that it is hard for an automated technique to distinguish among the \objectname, \actorname, and \passivename\ roles, when there is no contextual information, since their values all correspond to nouns. For instance, a ``\textit{user}'' may be tagged as \objectname\ rather than \actorname, given that business objects are much more common in the training data  and there is no context that indicates the correct role. 
Therefore, we establish a set $\textattributes_n \subseteq \textattributes$ that contains all such \emph{noun-only} attributes, i.e. attributes of which all values correspond solely to the \objectname\ role. This set is then forwarded to Step~3, whereas the tagged values of the other attributes directly become part of our approach's output.

\subsection{Step 3: Attribute-level classification}
\label{sec:approachstep3}

In this step, the approach determines the semantic role of miscellaneous attributes, $\miscattributes$ identified in Step~1, and the noun-only textual attributes, $\textattributes_n$, identified in Step~2. 
We target this at the attribute level, i.e., we determine a single semantic role for each $D \in \miscattributes \cup \textattributes_n$ and assign that role to each occurrence of $D$ in the event log. For attributes in $\miscattributes$, the approach determines the appropriate role (if any) based on an attribute's name, whereas for attributes in $\textattributes_n$, it considers the name as well as its values. 
Note that we initially assign each attribute a role $r \in \mathcal{R}'$, where $\mathcal{R}'$ excludes the \emph{instance} resource roles, i.e. \actorinstance\ and \passiveinstance,  and later distinguish between type-level and instance-level based on the attribute’s domain.

\mypar{Classifying miscellaneous attributes} 
To determine the role of miscellaneous attributes, we recognize that their values, typically alphanumeric identifiers, integers or Booleans, are mostly uninformative. Therefore, we determine the role of an attribute $D \in \miscattributes$ based on its name. In particular, we build a classifier that compares a $name(D)$ to a set of manually labeled attributes $\classifierattributes$, derived from real-life event logs $\mathcal{L}$ (with $L \notin \mathcal{L}$).

Using $\classifierattributes$, we built a multi-class text classifier function $\textit{classify}(D)$ that, given an attribute $D$, returns $r_{D} \in \labelroles  \cup \{\texttt{other}\}$ as the semantic role closest to $name(D)$, with $conf(r_D) \in [0, 1]$ as the confidence.
To this end, we encode the names from $\classifierattributes$ using the GloVe~\cite{pennington2014glove} vector representation for words. Subsequently, we train a logistic regression classifier on the obtained vectors, which can then be used to classify unseen attribute names. 
Since 
GloVe provides a state-of-the-art representation to detect \emph{semantic similarity} between words, the classifier can recognize that, e.g., an \texttt{item} attribute is more similar to \objectname\ attributes like \texttt{product} than to \actorname\ attributes in $\classifierattributes$.

\mypar{Classifying noun-only attributes}
Given an attribute in $D \in \textattributes_n$, we first apply the same classifier as used for miscellaneous attributes. If $\textit{classify}(D)$ provides a classification with a high confidence value, i.e., $conf(r_D) \geq \tau$ for a threshold $\tau$, our approach uses $r_D$ as the role for $D$. 
In this way, we directly recognize cases where $\textnormal{name}(D)$ is equal or highly similar to some of the known attributes in $\classifierattributes$. 
However, if the classifier does not yield a confident result, we instead analyze the textual values in $\textnormal{dom}(D)$.

\begin{figure}[!hbt]
	\centering
	\vspace{-1em}
	\includegraphics[width=\linewidth]{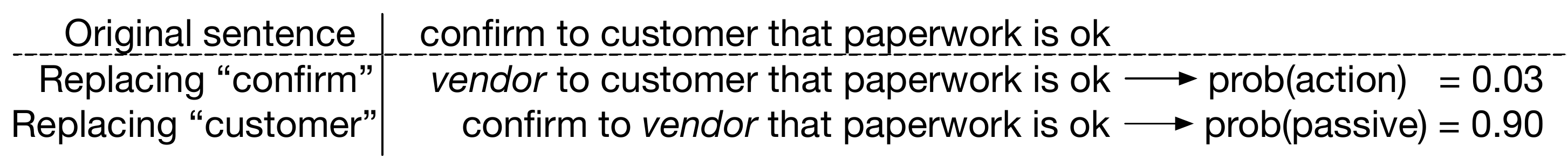}
	\caption{Exemplary insertion of a value from an attribute in $\textattributes_n$ into an existing context.}
	\label{fig:bert_insertion}
	\vspace{-1em}
\end{figure}

Since noun-only attributes were previously re-assigned due to their lack of context, we here analyze them by artificially placing each attribute value into contexts that correspond to different semantic roles.
In particular, as shown in \autoref{fig:bert_insertion}, we insert a candidate value (e.g., ``\textit{vendor}'') into different positions of a set $T$ of highly expressive textual attribute values (i.e., ones with at least 3 semantic roles). 
The resulting texts are then fed into the language model employed in Step~2, 
allowing our approach to recognize which context and, therefore, which semantic role, best suits the candidate value (i.e., \passivename\ in \autoref{fig:bert_insertion}). Finally,  we assign $r_D \in   \labelroles \cup \{\texttt{other}\}$ as the role that received the most votes across the different texts in $T$ and  values in $\textnormal{dom}(D)$.

\mypar{Recognizing instance-level attributes}
Since we only focused on the type-level roles $\mathcal{R}'$ in the above, we lastly check for every resource-related attribute $D \in \miscattributes$, with $r_D \in \{ \actorname, \passivename\}$, if it actually corresponds to an instance-level role instead. 
Particularly, we change $r_D$ to the corresponding instance-level role if $\textnormal{dom}(D)$ has values that contain a numeric part or only consist of named-entities (e.g., ``\emph{Pete}''). For instance, an attribute $D_1$ with values like \emph{user\_019} and \emph{batch\_06}, contains numeric parts and is, thus reassigned to \actorinstance, while an attribute $D_2$ with  $\textnormal{dom}(D_2$)$ = \{\textit{staff member, system}\}$  will retain its $\actorname$ role.

\subsection{Output}
\label{sec:approachoutput}

Given an event $e$, our approach returns a collection of tuples $(r, v)$ with $r \in \mathcal{R}$ a semantic role and $v$ a value, where $v$ 
either corresponds to an entire attribute value $e.D$ (for attribute-level classification applied to attributes in $\miscattributes \cup \textattributes_n$) or to a part thereof (stemming from the instance-level labeling applied to $\textattributes \setminus \textattributes_n)$.

To enable the  subsequent application of process mining techniques, the approach returns an XES event log $L'$ that contains these labels as additional event attributes, i.e., it does not override the names or values of existing ones. 
Note that we support different ways to handle cases where an event has multiple tuples
with the same semantic role, e.g., the ``\emph{draft}'' and ``\emph{send}'' actions stemming from a ``\textit{draft and send request}'' label: the values are either collected into one attribute, 
i.e.,  \actionname = [\emph{draft, send}], or into multiple, uniquely-labeled attributes, i.e., \actionname \texttt{:0} = \emph{draft},  \actionname \texttt{:1} = \emph{send}.
Furthermore, if multiple \objectstatus\ (or \actionstatus) attributes exist that each have Boolean values, e.g., \texttt{isCancelled} and \texttt{isClosed} for the Hospital log, these are consolidated into a single  attribute, for which events are assigned a value based on their original Boolean attributes, e.g., $\{\bot, isCancelled, isClosed\}$.

\section{Evaluation}
\label{sec:evaluation}

We implemented our approach as a Python prototype\footnote{\url{https://gitlab.uni-mannheim.de/processanalytics/extracting-semantic-process-information}}, using the PM4Py library~\cite{pm4py} for event log handling. 
Based on this prototype, we evaluated the accuracy of our approach and individual steps on a collection of 14 real-life event logs.

\subsection{Evaluation Data}
\label{sec:evaluationdata}

To conduct our evaluation, we selected all real-life event logs publicly available in the common 4TU repository\footnote{\url{https://data.4tu.nl/search?q=:keyword:\%20\%22real\%20life\%20event\%20logs\%22}}, except from those capturing data on software interactions or sensor readings, given their lack of natural language content. For collections that included multiple event logs with highly similar attributes, i.e., BPI13, BPI14, BPI15 and BPI20, we only selected one log per collection, to maintain objectivity of the obtained results. 
\autoref{tab:evallogs} depicts the details on the resulting collection of 14 event logs. 
They cover processes of different domains, for instance financial services, public administration and healthcare. Moreover, they vary significantly in their number of event classes, textual attributes, and miscellaneous attributes.

\begin{table}[!htb]
	\centering
	\caption{Characteristics of the considered event logs, with $\mathbfcal{C}$ as the set of event classes}
	\setlength{\tabcolsep}{4pt}
	\label{tab:evallogs}
		\begin{tabular}{llrcc l llrcc}
			\toprule
			\textbf{ID} & \textbf{Log name}  & $|\mathbfcal{C}|$& $|\mathbfcal{D}|$ & $|\mathbfcal{D}^{\mathbf{T}}|$&
			 \hphantom{x} & 
			\textbf{ID} & \textbf{Log name}  & $|\mathbfcal{C}|$& $|\mathbfcal{D}|$ & $|\mathbfcal{D}^{\mathbf{T}}|$ \\
			\midrule
			$L1$ & BPI12 & 24 & \hphantom{1}4  & 2 &  &  	$L8$ &BPI20	 & 51 & \hphantom{1}5 & 4 \\
			$L2$ & BPI13 & 4 & 11  & 4 &  &  						 $L9$ &CCC19  & 29  & 11 & 4 \\
			$L3$ &BPI14	 & 39  & \hphantom{1}5  & 2 &  & 	$L10$ &Credit Req. & 8 & \hphantom{1}4 & 3 \\
			$L4$ &BPI15	 & 289 & 13  & 3 &  & 					   $L11$ &Hospital  & 18 & 22 & 2 \\
			$L5$ &BPI17	 & 26 & 13 & 4 &  &  						$L12$ &RTFM & 11 & 15 & 2 \\
			$L6$ &BPI18	 & 41 & 13  & 5 &  & 						$L13$ &Sepsis   & 16  & 31 & 1 \\
			$L7$ &BPI19	 & 42 & \hphantom{1}4  & 2 &  & 	$L14$ &WABO  & 27 & 6 & 2 \\
			\bottomrule
		\end{tabular}
\end{table}

\subsection{Setup}
As a basis for our evaluation, we jointly established a \emph{gold standard} in which we manually annotated all unique textual values (for instance-level labeling) and attributes (for attribute-level classification) with their proper semantic roles\footnote{For reproducibility, the gold standard is published alongside the implementation.}.
Since our approach requires training for the language model used in the instance-level labeling (\autoref{sec:approachstep2}) and for the attribute-name classifier (\autoref{sec:approachstep3}), 
we perform our evaluation experiments using \emph{leave-one-out} cross-validation, in which we repeatedly train our approach on 13 event logs and evaluate it on the 14th.  This procedure is repeated such that each log in the collection is considered as the test log once.

To assess the performance of our approach, we compare the annotations obtained using our approach against the manually created ones from the gold standard. Specifically, we report on the standard \emph{precision}, \emph{recall}, and the \emph{F$_1$-score}.
Note that for instance-level labeling, we evaluate correctness per chunk, e.g., if a chunk (\emph{purchase order}, \objectname) is included in the gold standard, both ``\emph{purchase}'' and ``\emph{order}'' need to be associated with the \objectname\ role in the result, otherwise, neither is considered correct.

\subsection{Results}
\autoref{tab:results} provides an overview of the main results of our evaluation experiments. In the following, we first consider the performance of the instance-level labeling and attribute-level classification steps separately, before discussing the overall performance.
\begin{table}[!htb]
	\centering
	\caption{Results of the evaluation experiments}
	\label{tab:results}
	\begin{tabular}{l cccc @{\hskip 1em}cccc @{\hskip 1em}cccc}
		\toprule
		& \multicolumn{4}{c}{\textbf{Instance-level}} & 		 \multicolumn{4}{c}{\textbf{Attribute-level}} & 
		\multicolumn{4}{c}{\textbf{Overall}} \\ 
		\textbf{Semantic role} &   
		\textbf{Count} & \textbf{Prec.} & \textbf{Rec.} & $\mathbf{F_1}$ & 
		\textbf{Count} & \textbf{Prec.} & \textbf{Rec.} & $\mathbf{F_1}$ & 
		\textbf{Count} & \textbf{Prec.} & \textbf{Rec.} & $\mathbf{F_1}$ 
		\\
		\midrule
		\objectname      & \hphantom{1}583 & 0.89  & 0.88 & 0.88  & \hphantom{1}2    & 0.50 & 0.50 & 0.50 & \hphantom{1}585 & 0.89 & 0.88 & 0.88\\
		\objectstatus    & \hphantom{15}31    & 0.85  & 0.77  & 0.78  & \hphantom{1}6   & 0.50 & 0.33 & 0.40 &\hphantom{15}37 &  0.79 & 0.70 & 0.72\\
		\actionname      & \hphantom{1}630 & 0.94  & 0.95  & 0.94 & \hphantom{1}0   & -       & -      & -      & \hphantom{1}630 & 0.94 & 0.95 & 0.94 \\
		\actionstatus    & \hphantom{15}27    & 0.85  & 0.81  & 0.82 & \hphantom{1}6   & 1.00 & 1.00 & 1.00  & \hphantom{15}33 & 0.88  & 0.84 & 0.85\\
		\actorname       & \hphantom{15}69    & 0.93  & 0.84 & 0.88  & \hphantom{1}0   & -      & -       & -      & \hphantom{15}69 & 0.93 & 0.84 & 0.88\\
		\actorinstance  & \hphantom{150}0      & -        & -       & -       & 					16 & 1.00 & 0.94 & 0.97 & \hphantom{15}16& 1.00 & 0.94 & 0.97 \\
		\passivename   & \hphantom{15}19     & 0.84  &  1.00 & 0.91  & \hphantom{1}0   & -      & -       & -      & \hphantom{15}19 & 0.84  & 1.00 & 0.91\\
		\midrule 
		Overall & 1,359      & 0.91       & 0.91       & 0.91       & \hphantom{1}30   & 0.87 & 0.79 & 0.83 & 1,389 & 0.91 & 0.90 & 0.90\\
		\bottomrule
	\end{tabular}
\end{table}

\mypar{Instance-level labeling results}
The table reveals that our instance-level labeling approach is able to detect semantic roles in textual attributes with high accuracy, achieving an overall $F_1$-score of 0.91. 
The comparable precision and recall scores, e.g. 0.94 and 0.95 for \actionname\ or 0.89 and 0.88 for \objectname, each suggest that the approach can accurately label roles while avoiding false positives.
This is particularly relevant, given that nearly half of the textual attribute values also contain information beyond the scope of the semantic roles considered here (see also \autoref{tab:training}).
An in-depth look reveals that the approach even performs well on complex values, such as ``\textit{t13 adjust document x request unlicensed}''. It correctly recognized the business objects (\emph{document} and \emph{request}), the action (\emph{adjust}) and status (\emph{unlicensed}), omitting the superfluous content (\emph{t13} and \emph{x}).

\mypartwo{Challenges}
We observe that the primary challenge for our approach relates to the differentiation between relatively similar semantic roles, namely between the two kinds of statuses, \objectstatus\ and \actionstatus, as well as the two kinds of resources, \actorname\ and \passivename. 
Making this distinction is particularly difficult in cases that lack sufficient contextual information or proper grammar. For example, an attribute value like ``\emph{denied}'' can refer to either type of status, whereas it is even hard for a human to determine whether the ``\textit{create suspension competent authority}'' label describes \emph{competent authority} as a primary actor or a passive resource.


\mypartwo{Baseline comparison}
To put the performance of our approach into context, we also compared its instance-level labeling step to a baseline: a state-of-the-art technique for the parsing of process model activity labels by Leopold et al.~\cite{Leopold2019}. For a fair comparison, we retrained our approach on the same training data as used to train the baseline (corresponding to the collection of process models in \autoref{tab:training}) and only assess the performance with respect to the recognition of \emph{business objects} and \emph{actions}, since the baseline only targets these. \autoref{tab:compperf} presents the results obtained in this manner for the event labels from all 14 considered event logs. 

The table shows that our approach greatly outperforms the baseline, achieving an overall $F_1$-score of 0.75 versus the baseline's 0.47. Post-hoc analysis reveals that this improved performance primarily stems from event labels that are more complex (e.g., multiple actions, various semantic roles or compound nouns spanning multiple words) or lack a proper grammatical structure. This is in line with expectations, given that the baseline approach has been developed to recognize several established labeling styles, whereas we observe that event data often does not follow such expectations. Finally, it is worth observing that the performance of our approach in this scenario is considerably lower than when trained on the full data collection (e.g., an $F_1$ of 0.66 versus 0.88 for the \objectname\ role), which highlights the benefits of our data augmentation strategies. 


\begin{table}[!htb]
	\centering
	\caption{Comparison of our instance-level labeling approach against a state-of-the-art label parser; both trained on process model activity labels and evaluated on event labels.} 
	\label{tab:compperf}
	\begin{tabular}{l c @{\hskip 1em}lll @{\hskip 1em}lll}
		\toprule
			& & \multicolumn{3}{l}{\textbf{Our approach}} & \multicolumn{3}{c}{\textbf{Baseline}~\cite{Leopold2019}} \\ 
			\textbf{Semantic role} & \textbf{Count} & \textbf{Prec.} & \textbf{Rec.} & $\mathbf{F_1}$ & \textbf{Prec.} & \textbf{Rec.} & $\mathbf{F_1}$ \\
			\midrule
			\objectname	& \hphantom{1}562 & 0.65 &0.68 & 0.66 &  0.40 & 0.40 & 0.40  \\
			\actionname	&  \hphantom{1}618 & 0.86 & 0.81 & 0.83 &  0.59 & 0.48 & 0.53  \\
			\midrule 
			Overall & 1,180 & 0.76 & 0.75& 0.75 & 0.50& 0.44 & 0.47\\
			\bottomrule
		\end{tabular}
\end{table}

\mypar{Attribute-level classification results}
As shown in \autoref{tab:results}, our also approach achieves good results on the attribute-level classification of attributes, with an overall precision of 0.87, recall of 0.79, and an $F_1$ of 0.83.
We remark that the outstanding performance of our approach with respect to the \actionstatus\ and \actorinstance\ roles is partially due to the usage of standardized XES names for some of these attributes, enabling easy recognition. Yet this is not always the case. For instance, 7 out of 16 \actorinstance\ attributes handled by this step use alternatives to the XES standard, such as \texttt{User} or \texttt{Assingment\_Group}. 
Our approach maintains a high accuracy for these cases, correctly recognizing 6 out of 7 of such attributes.
Notably, the overall precision of our attribute-classification technique reveals that it is able to avoid false positives well, even though a substantial amount of event attributes are beyond the scope of our semantic roles, such as monetary amounts or timestamps. This achievement can largely be attributed to the domain analysis employed in our approach's first step.

Nevertheless, it is important to consider that these results were obtained for a relatively small set of 30 non-textual attributes. 
 Therefore, the lower results for certain uncommon semantic roles (e.g., \objectname), as well as the overall high accuracy for this step should be considered with care. This caveat also highlights the need additional training data, in order to expand the generalization of this part of our approach.

\mypar{Overall results}
The overall performance of the approach can be considered as the average
over the instance-level and attribute-level results, weighted against the number of entities that were annotated (cf., \emph{count} in \autoref{tab:results}), i.e., a unique textual attribute value (instance-level) or an entire attribute (attribute-level). 

We observe that the approach achieves highly accurate overall results, with a micro-average precision of 0.91, and a recall and $F_1$-score of 0.90. Still, when considering the results per semantic role, we observe that there exist considerable differences. These differences are largely due to the lower scores obtained for the underrepresented roles in the data set, since it is clear that our approach is highly accurate on more common roles, such as the $F_1$ score of 0.94 for the recognition of actions.  

\section{Case Study}
\label{sec:application}

This section demonstrates some of the benefits to be obtained by using the semantic information extracted by our proposed approach. To this end, we applied our approach to the \textit{Permit Log} published as part of the BPI20 collection~\cite{bpi20}, which contains 7,065 cases and 86,581 events, divided over 51 event classes (according to the event label, i.e., the \texttt{concept:name} attribute).
By applying our approach on the log, we identify information on five semantic roles. 
Most prominently, our approach is able to  extract information about the \actionname, \actionstatus, \objectname, and \actorname\ roles from the log's unstructured, textual event labels. 
The availability of these semantic roles as attributes in the augmented event log, created by our approach, enables novel analyses, such as:

\mypar{Event class refinement}
The event log contains event labels  that are polluted with superfluous information, e.g., by including resource information such as  `\emph{by budget owner}', resulting in a total of 51 event classes.
Any process model derived on the basis of these classes, therefore, automatically exceeds the recommended maximum of 50 nodes in a process model~\cite{mendling2010seven}, which impedes its understandability. To alleviate this, we can use the output of our approach to refine the event classes  by
grouping together events that involve the same \actionname\ and \objectname. For instance, we group events with labels like  ``\textit{declaration approved by budget owner}'' and ``\textit{declaration approved by administration}'', while deferring the actor information to a dedicated \actorname\ attribute. 
In this manner, we reduce the number of event classes from 51 to 21, which yields smaller and hence more understandable process models through process discovery techniques.

\begin{figure}[!htbp]
	\centering
	\includegraphics[width=1\linewidth]{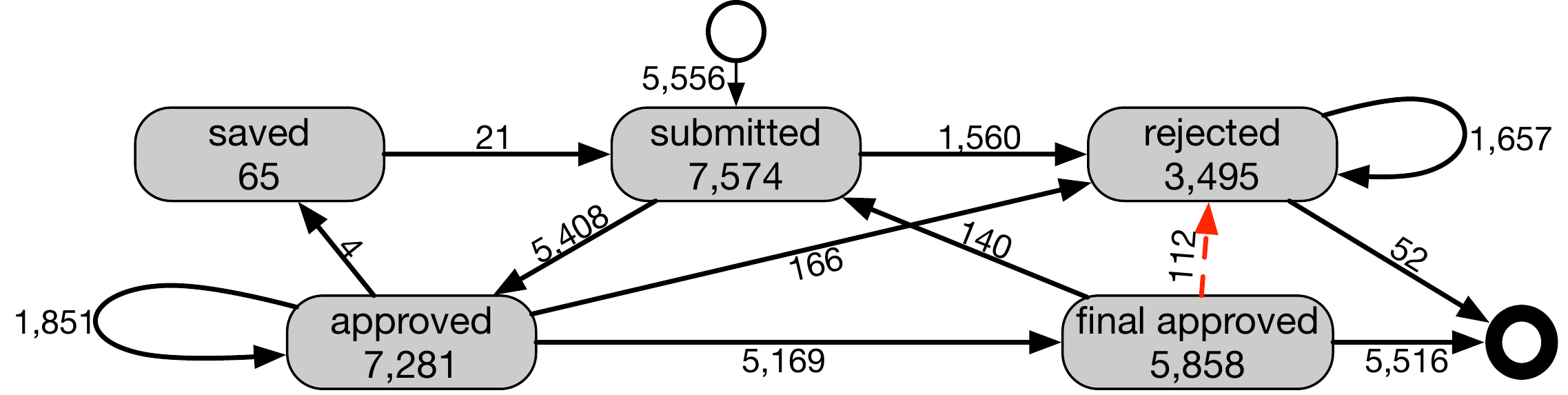}
	\caption{Example for object-centric analysis. The directly-follows graph shows the actions applied to the object \textit{declaration} in the log (includes 100\% activities, 50\% paths).}
	\label{fig:declaration_actions}
\end{figure}

\mypar{Object-centric analysis}
The extracted semantic information also enables us to investigate the behavior associated with specific business objects. Through the analysis of event labels, our approach recognizes that the  log contains six of these: \textit{permit}, \textit{trip}, \textit{request for payment}, \textit{payment}, \textit{reminder}, and \textit{declaration}. 
In \autoref{fig:declaration_actions} we show the directly-follows graph computed for the latter, obtained by selecting all events with $e.\objectname = `declaration'$, and using the identified actions to establish the event class. 
The figure clearly reveals how declarations are handled the process. Mostly, declarations are \emph{submitted}, \emph{approved}, and then \emph{final approved}. Interestingly, though, we also see 112 cases in which a declaration was definitely approved, yet rejected afterwards. 

It is important to stress that both the \textit{event class refinement} and \textit{object-centric analysis} are based on information extracted from the unstructured, textual labels of the \texttt{concept:name} attribute in the original log. Therefore, the presented insights cannot be obtained by manually categorizing the attributes of the event log, but rather require the thorough, instance-level event analysis provided by our approach.



\section{Related Work}
\label{sec:relatedwork}

Our work primarily relates to streams of research focused on the analysis of event and process model activity labels, as well as to the semantic role labeling task in NLP. 

Various approaches strive to either disambiguate or consolidate labels in event logs. 
Lu et al.~\cite{Lu2016} propose an approach to detect duplicate event labels, i.e., labels that are associated with events that occur in different contexts. By refining such duplicates, the quality of subsequently applied process discovery algorithms can be improved.  
Work by Sadeghianasl et al.~\cite{Sadeghianasl2020} aim to detect the opposite case, i.e., situations in which different labels are used to refer to behaviorally equivalent events.
Other approaches strive for the semantic analysis of labels, such as work by Deokar and Tao~\cite{Deokar2015}, which group together event classes with semantically similar labels, as well as the label parsing approach by Leopold et al.~\cite{Leopold2019} against which we compared our work in the evaluation. 
Finally, complementary to our approach, work by Tsoury et al.~\cite{tsoury2018conceptual} strives to augment logs with additional information derived from database records and transaction logs.

Beyond the scope of process mining, our work also relates to semantic annotation applied in various other contexts. Most prominently, semantic role labeling is a widely recognized task in NLP \cite{srl, jurafsky}, which labels spans of words in sentences that correspond to semantic roles. The tasks' goal is to answer questions like \emph{Who is doing what, where and to whom?} While early work in this area mostly applied feature engineering methods \cite{pradhan2005}, recently deep learning-based techniques have been successfully applied, e.g., \cite{he2017, zhang2020}. In the context of web mining, semantic annotation focuses on assigning semantic concepts to columns of web tables \cite{zhang2017}, while in the medical domain it is e.g. used to extract the symptoms and their status from clinical conversations \cite{du2019}.

\section{Conclusion}
\label{sec:conclusion}

In this paper, we proposed an approach to extract semantic information from events recorded in event logs. Namely,  it extracts up to eight semantic roles per event, covering business objects, actions, actors, and other resources, without imposing any assumptions on the structure of an event log's attributes. 
We demonstrated our approach's efficacy through evaluation experiments using a wide range of real-life event logs. The results show that our approach accurately extracts the targeted semantic roles from textual attributes, while considerably outperforming a state-of-the-art activity label parser in terms of both scope and accuracy, whereas our attribute classification techniques were also shown to yield satisfactory results when dealing with the information contained in non-textual attributes. 
Finally, we highlighted the potential of our work by illustrating some of its benefits in an application scenario based on real-life data. Particularly, we showed how our approach can be used to refine and consolidate event classes in the presence of polluted labels, as well as to obtain object-centric insights about a process.

In the future, we aim to expand our work in various directions. To improve its accuracy, we aim to  include data from external resources such as common sense knowledge graphs or dictionaries of domain-specific vocabulary into the approach. Furthermore, we intend to broaden its scope by introducing additional kinds of semantic roles, such as roles that disambiguate between human actors and systems.
However, most importantly, through its identification of semantic information, our work provides a foundation for the development of wholly novel, semantics-aware process mining techniques.

\smallskip
\noindent \textit{Reproducibility: The implementation, dataset, and gold standard employed in our work are all available through the repository linked in \autoref{sec:evaluation}.}

%
%
%
%
\bibliographystyle{splncs04}
\bibliography{refs}
\end{document}